\documentclass[twoside,11pt]{article}

\usepackage{listings}
\usepackage{graphicx} 
\usepackage{subcaption}
\usepackage{enumitem}
\usepackage{float}
\usepackage{jmlr2e}

\usepackage{color}
\definecolor{gray}{rgb}{0.4,0.4,0.4}
\definecolor{darkblue}{rgb}{0.0,0.0,0.6}
\definecolor{cyan}{rgb}{0.0,0.6,0.6}
\definecolor{maroon}{rgb}{0.5,0,0}
\definecolor{darkgreen}{rgb}{0,0.5,0}

\lstdefinelanguage{XML}
{
	basicstyle=\ttfamily\tiny,
	morestring=[s]{"}{"},
	morecomment=[s]{?}{?},
	morecomment=[s]{!--}{--},
	commentstyle=\color{darkgreen},
	moredelim=[s][\color{black}]{>}{<},
	moredelim=[s][\color{red}]{\ }{=},
	stringstyle=\color{blue},
	identifierstyle=\color{maroon},
	tabsize=3
}

\lstdefinelanguage{R}
{
	basicstyle=\ttfamily\tiny,
	keywords={t, library},
	otherkeywords={<<-,\%*\%},    %% cannot add <-
	comment=[l]{\#},
	keywordstyle=\color{blue},
	commentstyle=\color{darkgreen},
	tabsize=3
}

\lstdefinelanguage{Ruleset}
{
	basicstyle=\ttfamily\tiny,
	tabsize=3
}

\ShortHeadings{RuleKit}{Gudy\'{s}, Sikora, and Wr\'obel}
\firstpageno{1}

\begin{document}

\title{RuleKit: A Comprehensive Suite for Rule-Based Learning}

\author{\name Adam Gudy\'{s} \email adam.gudys@polsl.pl
       	\\
       	\name Marek Sikora \email marek.sikora@polsl.pl
        \\
       \name \L{}ukasz Wr\'obel \email lukasz.wrobel@polsl.pl \\ \\
      	\addr Institute of Informatics, Silesian University of Technology, 44-100 Gliwice, Poland}
\editor{}

\maketitle

\begin{abstract}%   <- trailing '%' for backward compatibility of .sty file
Rule-based models are often used for data analysis as they combine interpretability with predictive power. We present RuleKit, a versatile tool for rule learning. Based on a sequential covering induction algorithm, it is suitable for classification, regression, and survival problems. The presence of a user-guided induction facilitates verifying hypotheses concerning data dependencies which are expected or of interest. The powerful and flexible experimental environment allows straightforward investigation of different induction schemes. The analysis can be performed in batch mode, through RapidMiner plug-in, or R package. A documented Java API is also provided for convenience. The software is publicly available at GitHub under GNU AGPL-3.0 license.

\end{abstract}


\section{Introduction}
Thanks to the combination of predictive and descriptive capabilities, rules have been applied in machine learning (especially in knowledge discovery) for decades. Amongst many rule induction strategies, sequential covering is one of the most popular~\citep{furnkranz2012foundations}. It consists in iterative addition of rules explaining a part of the training set as long as all the examples are covered. This approach leads to different models than those obtained by extracting rules from trees induced with a divide-and-conquer strategy~\citep{BreimanFOS84}. In the previous research we confirmed the effectiveness of our variant of sequential covering strategy on dozens of data sets in classification, regression, and survival analysis~\citep{wrobel2016,wrobel2017}. We also showed a usefulness of user-guided induction, which allows introducing user's preferences or domain knowledge to the learning process~\citep{sikora2019}---feature particularly valuable in medical applications. 

In spite of numerous advantages, relatively few sequential covering rule induction algorithms are available as ready-to-use software. The examples are CN2~\citep{clark1989cn2} included in the Orange suite~\citep{demvsar2013orange}, AQ~\citep{michalski1969quasi} implemented in Rseslib~3~\citep{rseslib2019}, or RIPPER~\citep{cohen1995fast} and M5Rules~\citep{Holmes1999} contained in Weka~\citep{Witten2016}.

We present RuleKit, a comprehensive suite for training and evaluating rule-based data models. Equipped with multiple useful features like user-guided induction, it is the first tool suitable for classification, regression, and survival analysis problems. It additionally stands out from the competitors with handiness---beside batch experimental environment it can be integrated with RapidMiner and R. 

\section{RuleKit Features}
The following features make RuleKit a powerful data analysis tool:
\begin{enumerate}[noitemsep,topsep=0pt,parsep=0pt,partopsep=0pt]
	\item[(i)]{Ability to resolve different problems: classification, regression, and survival analysis.} 
	\item[(ii)]{Various ways to run the analysis: batch mode, RapidMiner plug-in, R package.}
	\item[(iii)]{Multiplicity of algorithm parameters. For instance, there are over 40 rule quality measures available with an additional possibility to define own formulas.}
	\item[(iv)]{Integrated experimental environment---the software facilitates automated investigation of various algorithm configurations over multiple data sets. Different experimental schemes (train-test, cross validation) are supported and tens of performance metrics are provided for model assessment.}
	\item[(v)]{User-guided induction---the possibility to specify the initial set of rules, preferred and forbidden conditions/attributes, together with the multiplicity of options and modes allow suiting the model to user's requirements. This may be useful, e.g., in verifying hypotheses concerning data dependencies which are expected or of interest.}
	\item[(vi)]{Computational scalability---independent steps of induction algorithms (e.g., the evaluation of different conditions) are distributed over multiple threads allowing RuleKit to take advantage of multi-core CPUs, as well as high-performance clusters. Bit-level parallelism is also employed for maximum computational performance.}
	\item[(vii)]{Portability---the suite is distributed as Java application, thus it can be run on the majority of operating systems, including Windows, Linux, and OS X.}
	\item[(vii)]{Extensibility---the software together with the source code is publicly available at GitHub under GNU AGPL-3.0 Licence: 
	\url{https://github.com/adaa-polsl/RuleKit}.
	The documented API allows straightforward integration of the library with other projects and/or extending its functionality.}

\end{enumerate} 

\section{Case Studies}
%\subsection{Batch mode}
\textbf{Batch mode.} 
This example demonstrates running a RuleKit batch analysis on \textit{deals} classification data set (prediction whether a person making a purchase will be a future customer). The batch mode is run with \texttt{java -jar RuleKit experiments.xml} command, where XML file describes parameter sets and data sets to be investigated (Figure~\ref{lst:batch-xml}\,a).

As a result of the training, a text report is produced (Figure~\ref{lst:batch-xml}\,b). It contains a list of generated rules (with corresponding confusion matrices and statistical significance), information about examples coverage, model characteristics (no. of rules/conditions, average rule precision/coverage, etc.), and performance metrics calculated on the training set (accuracy, error, etc.). Depending on the problem, the significance of rules is established with different tests (Fisher's exact, ${\chi}^2$, or log-rank). The training may be followed by applying the model on unlabelled data. In this stage, a comma-separated table with values of performance metrics evaluated on the test set is produced. 

\newpage

\begin{figure}[t]
	\begin{minipage}[t]{.49\textwidth}
		\begin{lstlisting}[language=XML] 
		<experiment>
			<parameter_sets>
			<parameter_set name="mincov=8, Entropy_User_C2">
				<param name="min_rule_covered">8</param>
				<param name="induction_measure">BinaryEntropy</param>
				<param name="pruning_measure">UserDefined</param>
				<param name="user_pruning_equation">2 * p / n</param>
				<param name="voting_measure">C2</param>
				</parameter_set>
			</parameter_sets>
			
			<datasets>
				<dataset>
					<label>Future Customer</label>
					<out_directory>./results-deals</out_directory>		
					<training>  
						<report_file>training.log</report_file>           		
						<train>
						<in_file>../data/deals/deals-train.arff</in_file>               	
						<model_file>deals.mdl</model_file> 
						</train>
					</training>
					<prediction>
						<performance_file>performance.csv</performance_file>  
						<predict>
							<model_file>deals.mdl</model_file>      	
							<test_file>../deals/data/deals-test.arff</test_file>            			
							<predictions_file>deals-pred.arff</predictions_file>  	  
						</predict>
					</prediction>
				</dataset>
			</datasets>
		</experiment>
		\end{lstlisting}
		\subcaption{}		
	\end{minipage}\hfill
	\begin{minipage}[t]{.49\textwidth}
		
		\begin{lstlisting}[language=Ruleset] 
		Rules:
		r1: IF Gender = {male} AND Age = (-inf, 34.5) 
		THEN Future Customer = {yes} 
		(p=176.0, n=0.0, P=473.0, N=527.0, weight=0.69, pval=3.8E-67)
		r2: IF Payment Method = {credit card} AND Age = (-inf, 30.5) 
		THEN Future Customer = {yes} 
		(p=183.0, n=0.0, P=473.0, N=527.0, weight=0.69, pval=2.9E-70)
		r3: IF Gender = {male} AND Age = (-inf, 35.5) 
		THEN Future Customer = {yes} 
		(p=185.0, n=0.0, P=473.0, N=527.0, weight=0.70, pval=3.6E-71)
		...	
		
		Best rules covering examples from training set (1-based):
		5*,12;3*,12;2*,14,15;6,11*,12,13,14,15;5*,12;13*,14,15;2*,15;...
		
		Model characteristics:
		time_total_s: 0.916438051
		time_growing_s: 0.684499623
		time_pruning_s: 0.365774315
		#rules: 15.0
		#conditions_per_rule: 2.0
		#induced_conditions_per_rule: 22.533333333333335
		avg_rule_coverage: 0.26126666666666665
		avg_rule_precision: 0.9517768845699609
		...
		
		Training set performance:
		accuracy: 0.954
		classification_error: 0.04600000000000004
		kappa: 0.9072689080712335
		balanced_accuracy: 0.9513742071881607
		#rules_per_example: 3.919
		...
		
		\end{lstlisting}
		\subcaption{}			
	\end{minipage}
	\caption{(a) XML batch experiment on \textit{deals} data set. (b) Resulting training report.}\label{lst:batch-xml}
\end{figure}

\begin{figure}[t]
	\begin{subfigure}[b]{0.56\textwidth}
		\includegraphics[width=\textwidth]{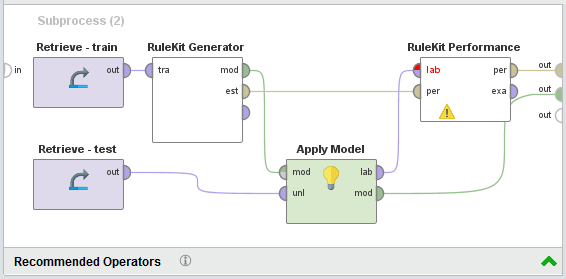}
		\caption{}
	\end{subfigure}
	~ %add desired spacing between images, e. g. ~, \quad, \qquad, \hfill etc. 
	\begin{subfigure}[b]{0.41\textwidth}
		\includegraphics[width=\textwidth]{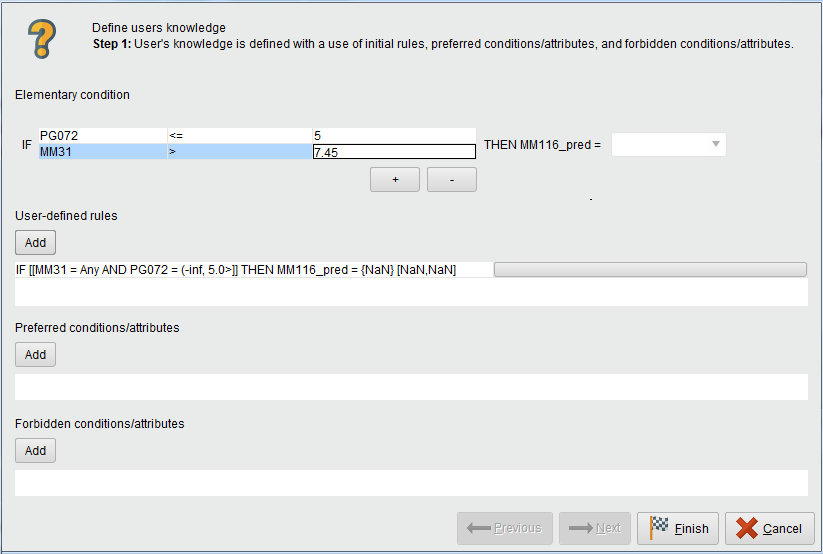}
		\caption{}
	\end{subfigure}
	\caption{(a) Process for analyzing \textit{methane} data set with RuleKit RapidMiner plug-in.\\ (b) Wizard for specifying user's knowledge in the guided induction.}\label{fig:plugin}
\end{figure} 
%
%\subsection{RapidMiner Plug-in}
\noindent
\textbf{RapidMiner plug-in.}
An alternative way of performing an experiment is integrating RuleKit with RapidMiner. The plug-in provides user with two operators: \textit{RuleKit Generator} and \textit{RuleKit Performance}. The former is a RapidMiner \textit{learner} that induces various types of rule models. The latter extends the standard RM \textit{Performance} operator and allows calculation of performance metrics as well as gathering model characteristics. In the Figure~\ref{fig:plugin} we present an example RapidMiner process which performs regression analysis on the \textit{methane} data set (predicting methane concentration in a coal mine) and a wizard for specifying user's knowledge in the guided induction.

\begin{figure}[t]
	\begin{minipage}[b]{.49\textwidth}
		\begin{lstlisting}[language=R] 
		library(ggplot2)
		library(reshape2)
		library(rulekit)
				
		formula <- survival::Surv(survival_time,survival_status) ~ .
		control <- list(min_rule_covered = 5)
		results <- rulekit::learn_rules(formula,control,bone_marrow)
		
		# extract outputs:
		rules <- results[["rules"]] # list of rules
		cov <- results[["train-coverage"]] # coverage information
		surv <- results[["estimator"]] # survival function estimates
		perf <- results[["test-performance"]]  # testing performance
		
		# melt data set for automatic plotting of multiple series
		melted_surv <- reshape2::melt(surv, id.var="time")
		
		# plot survival functions estimates
		ggplot(melted_surv, aes(x=time, y=value, color=variable)) +
			geom_line(size=1.0) +
			xlab("time") + ylab("survival probability") +
			theme_bw() + theme(legend.title=element_blank())
		
		\end{lstlisting}
		\subcaption{}	
	\end{minipage}
	\begin{minipage}[b]{.49\textwidth}
		\begin{subfigure}[]{\textwidth}
		\includegraphics[width=\textwidth]{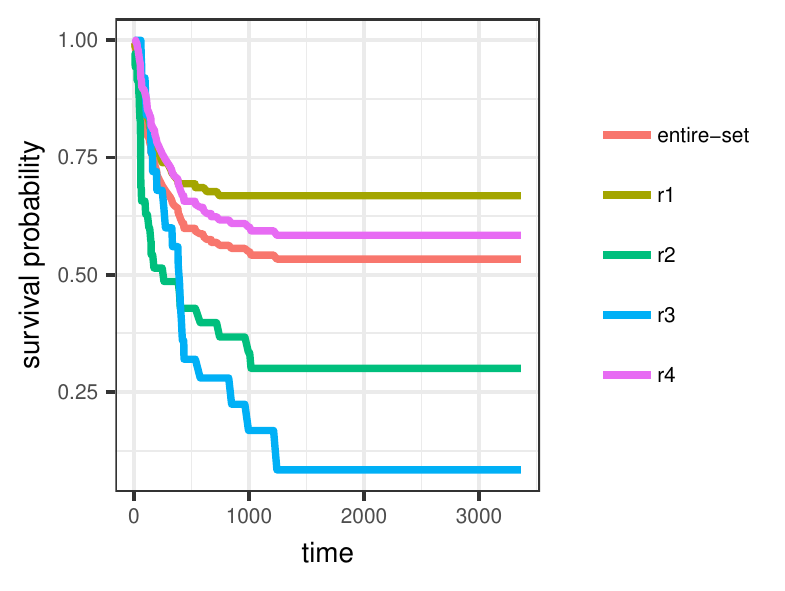}
		\end{subfigure}
		\subcaption{}			
	\end{minipage}
	\caption{Analyzing \textit{BMT-Ch} survival data set using RuleKit R package. (a) The code for training and visualizing model, (b) survival estimates of generated rules.}\label{lst:r-package}
\end{figure}

\noindent
%\subsection{R Package}
\textbf{R package.}
As a last test case, we present the application of RuleKit R package for analyzing factors contributing to the patients' survival following bone marrow transplants. The corresponding data set (\textit{BMT-Ch}) is integrated with the package in the form of the standard R data frame. The training and applying a model is performed by a function \texttt{learn\_rules} which returns a named list containing induced rules, survival function estimates, test set performance metrics, etc. In Figure~\ref{lst:r-package} we provide an example R code for training the model and visualizing corresponding survival functions estimates.

\section{Conclusions and Future Work}
We demonstrated that RuleKit can be successfully applied for training and evaluation of rule-based models in classification, regression, and survival tasks. The multiplicity of options and modes together with the powerful and flexible experimental environment makes presented suite a useful tool for data analysis and knowledge discovery. In the future, we plan to extend RuleKit with algorithms for inducing action rules~\citep{hajja} and oblique rules~\citep{sikora2013chira}. The applicability of the suite could be additionally enhanced by providing Python wrapper or standalone graphical interface.

% Acknowledgements should go at the end, before appendices and references

\acks{This work was supported by Polish National Centre for Research and Development (NCBiR) within the Operational Programme Intelligent Development (grant no. POIR.04.01.02-00-0024/17-00); Rector of Silesian University
of Technology (grant no. 02/020/RGJ18/0126); Institute of
	Informatics at Silesian University of Technology within the statutory
	research project (BKM18/RAU2/556). }


\newpage

%\appendix

%\vskip 0.2in
\bibliography{rulekit}

\end{document}